\documentclass[runningheads]{llncs}
\usepackage{todonotes}
\usepackage{graphicx}
\usepackage{hyperref}
\usepackage{amssymb}
\usepackage{amsmath}
\usepackage{mathtools}
\usepackage{trfrac}
\usepackage{listings}
\usepackage{cite}
\usepackage{xcolor}
\usepackage[]{cleveref}
\crefname{listing}{alg.}{algs.}  
\Crefname{listing}{Alg.}{Algs.}
\usepackage{subcaption}
\usepackage{import}
\usepackage{xifthen}
\usepackage{pdfpages}
\usepackage{transparent}
\usepackage{qtree}
\usepackage{booktabs}
\usepackage{caption}
\usepackage{csquotes}
\usepackage{float}

\newsavebox{\imagebox}

\DeclareMathOperator*{\true}{{\footnotesize\texttt{true}}}
\DeclareMathOperator*{\primif}{{\footnotesize\texttt{if}}}
\DeclareMathOperator*{\sign}{{\footnotesize\texttt{sign}}}
\DeclareMathOperator*{\Float}{{\footnotesize\texttt{Float}}}
\DeclareMathOperator*{\Bool}{{\footnotesize\texttt{Bool}}}
\DeclareMathOperator*{\T0}{{\footnotesize\texttt{T0}}}

\newcounter{nalg} % defines algorithm counter
\renewcommand{\thenalg}{\arabic{nalg}} %defines appearance of the algorithm counter
\DeclareCaptionLabelFormat{algocaption}{Algorithm \thenalg} % defines a new caption label as Algorithm x.y

\definecolor{cadet}{rgb}{0.33, 0.41, 0.47}

\lstnewenvironment{algorithm}[1][] %defines the algorithm listing environment
{   
    \refstepcounter{nalg} %increments algorithm number
    \captionsetup{labelformat=algocaption,labelsep=colon} %defines the caption setup for: it ises label format as the declared caption label above and makes label and caption text to be separated by a ':'
    \lstset{ %this is the stype
        mathescape=true,
        frame=tB,
        numbers=left, 
        numberstyle=\tiny,
        basicstyle=\scriptsize\ttfamily, 
        keywordstyle=\color{black}\bfseries,
        keywords={,optional, input, output, return, datatype, function, if, else, for, foreach, begin, end, } %add the keywords you want, or load a language as Rubens explains in his comment above.
        numbers=left,
        xleftmargin=.04\textwidth,
        morecomment=[l][\color{cadet}]{//},
        #1 % this is to add specific settings to an usage of this environment (for instnce, the caption and referable label)
    }
}
{}

\newcommand\Visiblespace[1][.7em]{%
  \mbox{\kern.06em\vrule height.5ex}%
  \vbox{\hrule width#1}%
  \hbox{\vrule height.5ex}}

\definecolor{codegreen}{rgb}{0,0.6,0}
\definecolor{codegray}{rgb}{0.3,0.3,0.3}
\definecolor{codepurple}{rgb}{0.58,0,0.82}
\definecolor{backcolour}{rgb}{0.95,0.95,0.92}

\title{Programmatic Policy Extraction by Iterative Local Search}

\author{Rasmus Larsen\inst{1}\orcidID{0000-0001-9177-2695} \and
Mikkel {Nørgaard Schmidt}\inst{1}\orcidID{0000-0001-6927-8869}}

\authorrunning{R. Larsen \and M. N. Schmidt}

\institute{Department of Applied Mathematics and Computer Science, \\
Technical University of Denmark, Kongens Lyngby, Denmark\\
\email{\{ralars,mnsc\}@dtu.dk}}

\begin{document}

\maketitle

\begin{abstract}
% Abstract
%  1. Motivation. Why do we care?
Reinforcement learning policies are often represented by neural networks, but programmatic policies are preferred in some cases because they are more interpretable, amenable to formal verification, or generalize better.
%  2. Problem formulation. What problem will we solve?
While efficient algorithms for learning neural policies exist, learning programmatic policies is challenging.
%     2.b. Current state. What have others done, and why is that not enough?
%  3. Approach. What is our big idea? How did we solve it?
Combining imitation-projection and dataset aggregation with a local search heuristic, we present a simple and direct approach to extracting a programmatic policy from a pretrained neural policy. 
%     3.a. Analysis and experiments. What research did we do?
%  4. Results. What is the answer?
After examining our local search heuristic on a programming by example problem, we demonstrate our programmatic policy extraction method on a pendulum swing-up problem.
%  5. Conclusions. What are the implications?
Both when trained using a hand crafted expert policy and a learned neural policy, our method discovers simple and interpretable policies that perform almost as well as the original. 
\end{abstract}

\section{Introduction}
% Introduction section
%  1. Introduction.
%Why programs
While neural policy representations are by far the most common in modern Reinforcement Learning (RL), other representations are worth considering. Programmatic policies provide a number of potential benefits: For example, a program might be read and understood by a human, something that generally is not possible with a neural network. Programs are also inherently compositional, which allows for not only reuse of policies in new combinations, but also compositional reasoning about their behavior.

%  2. Background and setting. 
%What's the issue
However, learning programmatic policies is challenging. The structured, discrete space of programs does not allow for the gradient based optimization that neural policies benefit greatly from. Compared to a more standard inductive synthesis setting, programmatic policies must be evaluated in an environment that, whether simulated or real, is expensive to interact with. Several approaches exist that attempt to handle this interaction issue, such as learning a parametric environment model \cite{Hein_Udluft_Runkler_2017}, imitating an existing policy \cite{Bastani_Pu_Solar-Lezama_2018, Verma_Murali_Singh_Kohli_Chaudhuri_2018}, or evaluating fewer programs by learning to search more efficiently \cite{Ellis2018}. Furthermore, \cite{Verma2019-ox} extend the imitation setting by providing a framework for intertwining RL and programmatic policy imitation. 

%Why use imitation projection instead.
This imitation-projection framework brings us a step closer to programmatic RL, where programs can be learned gradually through interaction with the environment. Essentially, this allows similar sample efficiency when compared to policy gradient methods, since the imitation-projection step is performed offline by scoring programs according to an imitation learning objective. One could even plausibly imagine that the inductive bias in a problem-specific policy language could lead to improved learning. 
The framework leaves many choices open in terms of how the policy update and programmatic policy projection steps are performed, as well as in terms of defining the program space. \cite{Verma2019-ox} perform experiments with a specific choice of update and projection, using two tailored program spaces based on PID controllers with either decision tree regression or Bayesian optimisation over some parameters as the projection operator.

%  3. Identification of problem

%  4. Purpose Statement. 
% What do we propose doing differently
In this paper we experiment with a more general program space based on Domain Specific Languages (DSLs) implemented in a typed lambda calculus. We demonstrate a method for re-using projections by local search around a previous projection, potentially reducing the required computational effort while allowing much longer programmatic policies to be found. Since imitation-projection greatly reduces environment interaction, the presented method takes advantage of this and performs relatively large searches in program space. Demonstrating the method on the pendulum swing-up task, we show that a simple and effective programmatic policy can be found by imitating a learned neural policy.

\section{Methods}
Our framework is based on previous work on imitation-projected and programmatically interpretable reinforcement learning \cite{Verma_Murali_Singh_Kohli_Chaudhuri_2018,Verma2019-ox}. We extend these methods to DSLs defined in a general-purpose programming language, namely the lambda calculus with Hindley-Milner type system. Using the building block of depth-limited type-directed program search, we construct an algorithm for finding a programmatic imitation of a given control policy. In order to discover programs much larger than what the depth limit of a single search allows, the algorithm performs multiple iterations of local search.

\subsection{Program synthesis by type-directed search}
\label{subsec:search}
% Program representation
By choosing the lambda calculus with Hindley-Milner type system, we obtain an expressive program space, in which exhaustive search for programs of a specified type is straightforwardly defined. An especially useful feature of this program representation is that the type system can be used to reduce the search space, by filtering candidate programs that do not type check. Further filtering is possible, such as the filtering of semantically similar programs as done in MagicHaskeller \cite{Katayama_2008}. In this program representation, DSLs can be defined as sets of typed functions and constants, which together represent the space of possible programs to be searched.

% Local search
Our starting point for program synthesis is a simple version of depth-limited type-directed search. This choice is not a given; other, more advanced program search algorithms can be used. Here, the space of all programs in a DSL is a tree, with the empty program at the root. Internal nodes are partial programs, with each branch being a candidate substitution for a hole in a partial program. Enumerating through this search tree results in generating all valid programs, according to the DSL, as the leaves of the tree. 

We take advantage of the typed language to reduce the search space. Instead of yielding all syntactically valid programs, as explained above, we want to yield only well-typed programs. This type-directed search algorithm is the same as used by e.g. \cite{Ellis2018} to sample programs from a prior distribution, but instead of sampling, all programs are enumerated. To expand a node in the search tree, a typed hole (an empty program with a type annotation) in the corresponding partial program is selected for synthesis. Then, valid candidates are selected from the set of all DSL candidates by unification; the resulting context of the unification is propagated to the further expansion of child nodes, ensuring that any constraints are satisfied. All candidates that can produce the correct type are considered, even if they would need arguments applied to them first.

\subsection{Typed neighborhood}

\begin{algorithm}[caption={Depth-limited local search (typed neighborhood)}, label={alg:search}, float=tp, floatplacement=tbp]
 input: domain specific language $\mathcal{D}$
 input: imitation dataset $\Gamma$
 input: initial program $P$
 output: best program in typed neighborhood $p^*$
 function $N_n^d(\mathcal{D},\, P,\, l) \quad$ // generates the neighborhood for location $l$ in $P$
   return $\varnothing$ if d = 0
   $T = \textsc{type}(P,\, l) \quad$  // type of expression at $l$ in $P$
   $C = \{e \,|\, e : t \in \mathcal{D} \land T \texttt{ can unify with } \textsc{yield}(t)\} \quad$  // everything valid from DSL
   $P' = \{\texttt{edit}(P, l, c)) \,|\, c \in C \}$
   // return all complete programs, recursively generate partial programs
   return $\{p \in P' \,|\, p \texttt{ is complete}\} \cup N_n^{d-1}(\mathcal{D}, p', l') \quad \forall p' \in \{p' \in P' \,|\, p' \texttt{ is partial}\}$
     where $l'$ is the location of the first hole in $p'$
 end
 $p^* \leftarrow \varnothing,\, v^* \leftarrow \infty \quad$  // best program and imitation loss
 foreach $l \in $ set of all paths in $P$
   $E_{l} = \textsc{expression}(P,\, l) \quad$ // expression at location 
   $\mathcal{D}' = \mathcal{D} \cup \left(E_{l},\, \textsc{type}(E_{l})\right) \quad$ // locally extend DSL with $E_l$
   foreach $p \in N_n^d(\mathcal{D}',\, P,\, l) \quad$
     evaluate $p$ on $\Gamma$ and update $p^*$ and $v^*$
   end
 end
\end{algorithm}

\begin{algorithm}[caption={Iterative local programmatic policy imitation}, label={alg:imitation}, float]
 input: oracle policy $f$ 
 optional input: initial program $p_{init} = \varnothing$
 output: imitation program $p_K$
 collect $N$ on-policy trajectories using $f$: 
   $\tau_0 = \left(\left(s^0_0, f(s^0_0), s^0_1, f(s^0_1), \dots\right), \dots, \left(s^N_0, f(s^N_0), s^N_1, f(s^N_1), \dots\right)\right)$
 create supervised dataset $\Gamma_0 = \left\{\left(s, f(s)\right) | s \in \tau_0 \right\}$
 derive $p_0$ from $\Gamma_0$ by local search from $p_{init} \quad$ // algorithm 1
 for $k = 1, \dots, K$
   collect $M$ on-policy trajectories using $p_{k-1}$:
     $\tau_k = \left(\left(s^0_0, p_{k-1}(s^0_0), s^0_1, p_{k-1}(s^0_1), \dots\right), \dots, \left(s^M_0, p_{k-1}(s^M_0), s^M_1, p_{k-1}(s^M_1), \dots\right)\right)$
   create supervised dataset $\Gamma' = \left\{\left(s, f(s)\right) | s \in \tau_k \right\}$
   aggregate datasets:
     $\Gamma_k = \Gamma_{k-1} \cup \Gamma' \quad$ // or $\Gamma_k = \Gamma_0 \cup \Gamma'$, which is cheaper
   derive $p_k$ from $\Gamma_k$ by local search from $p_{k-1} \quad$ // algorithm 1
 end
\end{algorithm}

%In this paper, the purpose of program synthesis is to approximate a policy, which is usually learned with reinforcement learning methods. In the setting considered, the policy is trained before program synthesis is performed, which is the PIRL setting \cite{Verma_Murali_Singh_Kohli_Chaudhuri_2018}. Despite that, a major consideration is how the program synthesis integrates in the iterative imitation-projection method. 
In \cref{alg:search}, the type-directed depth-limited synthesis algorithm is used to generate what we call the typed neighborhood of a given program.
This construction applies the basic synthesis algorithm in multiple places of an existing program, resulting in a iterative local search method that can both add and remove subprograms in each iteration. Because of this, the algorithm can synthesize larger programs, while also benefiting from work performed in previous iterations of the search. 

In more detail, we use a tree edit operation to define the programs contained within a typed neighborhood. Define the edit operation $\footnotesize\texttt{edit}(P,\, l,\, P')$ as the program obtained by replacing the subprogram at location $l$ in program $P$ with the program $P'$. Given a typed DSL $\mathcal{D}$, containing functions, constants, and their (polymorphic) types, the neighborhood of the program $P$ at location $l$ is the set of programs obtained by generating all well-typed expressions $P'$ contained in $\mathcal{D}$, written $N_n^d(\mathcal{D},\, P,\, l)$. The definition of a location is the root-to-expression path in the abstract syntax tree (AST) of the program. Here, we use the concept of a location in a generalized manner that can encompass multiple simultaneous locations, that is, a location $l$ can represent multiple paths in the AST that are to be simultaneously replaced using independent edit operations. The neighborhood of a program $P$ is thus the union of the neighborhoods at all locations, $N_n^d(\mathcal{D},\, P) = \bigcup_{l \in L(P)}N_n^d(\mathcal{D},\, P,\, l)$, where the neighborhood is parameterized with a maximum depth $d$ of the expressions generated by edit operations, and with a maximum number of simultaneous edits $n$.

Furthermore, the expression being edited is dynamically added as a candidate to the DSL, and for the depth evaluation this candidate counts as having a depth of 1. This allows an edit not just to replace an expression, but to also extend an expression by using it as part of the new expression, despite the result otherwise being too large (compared to the depth limit). The size of the neighborhood $|N^d_n(\mathcal{D}, P)|$ is quite sensitive to all involved parameters $\mathcal{D}$, $P$, $n$, and $d$, but these can be flexibly chosen based on the problem and available computational resources.

\subsection{Policy extraction by local synthesis}
We use the typed neighborhood to discover programs that imitate reinforcement learning policies. In this setting, input/output examples are obtained by executing an existing policy and storing the state observations together with the corresponding actions chosen by the policy in each state. Thus, the policy synthesis problem is framed as imitation learning. Like in previous policy imitation methods, an interactive dataset aggregation method such as DAgger \cite{Ross2010-zq} is used: Instead of imitating only on states that the expert experiences, which is called behavioral cloning, some experience from the imitation policy is periodically added to the set of states considered. This allows subsequent imitation iterations to correct mistakes that otherwise wouldn't be observed, since the expert policy never experiences these mistakes. However, unless the expert is a global optimum, it is possible that it also makes mistakes on states which are not usually observed, and for this reason it is not always a clear benefit. The iterative imitation approach using the typed neighborhood is described in \cref{alg:imitation}.
%Instead of fully training the RL policy before program synthesis, both RL and program synthesis steps are embedded in something akin to projected gradient descent. In the RL step, a standard method such as a policy gradient method \cite{Sutton_Barto_2018} is run for some number of steps in order to update the RL policy. In the program synthesis step, the RL policy is imitated, which in the optimal case can be seen as a projection of the policy onto the DSL. Practically, in most cases it is not possible to find the exact projection, i.e. the program $P = \argmin_{p \in \mathcal{D}^*}\mathcal{L}(p(x), y)$ where $\mathcal{D}^*$ is the set of all programs in a language $\mathcal{D}$, distance function $\mathcal{L}$, and imitation data $(x, y)$. Instead, the projection will only consider some tractable subset $\mathcal{D}_*$ of the programs $\mathcal{D}_* \subset \mathcal{D}^*$. \todo{notation here} It should however be considered that the exact projection of a neural policy, for example, is generally very complicated; thus, it might not be wanted for the sake of interpretability, and especially not for generalization purposes. Additionally, the \enquote{expert} policy is unlikely to be optimal everywhere in state space, and so obtaining an exact imitation does not have to be a benefit.

One purpose of this search algorithm is to fit into the full imitation-projection framework from \cite{Verma2019-ox}. A simple, modified version of this framework is shown in \cref{alg:ip}. The difference consists of the projection step, which now also depends on the previous projection, as enabled by \cref{alg:search}. Compared to \cref{alg:imitation}, the main difference is that each iteration also contains an {\footnotesize\texttt{Update}} step, which optimizes the expert policy.

\begin{algorithm}[caption={Imitation-Projected Programmatic Reinforcement Learning with Local Synthesis}, label={alg:ip}, float=tp, floatplacement=tbp]
 input: initial policy $\pi_0$
 optional input: initial program $p_0 = \varnothing$
 output: trained policy $\pi_J$, program $p_J$
 for $j = 1, \dots, J$
   $\pi_j \leftarrow \textsc{Update}(\pi_{j-1}) \quad$  // reinforcement learning, e.g. policy gradient
   $p_j \leftarrow \textsc{Project}(\pi_j,\, p_{init}=p_{j-1}) \quad$ // program synthesis by algorithm 2
 end
\end{algorithm}

\section{Experiments}
We present three different program synthesis experiments: The first is a programming by example (PBE) task with sampled ground truth programs, demonstrating the efficacy of the local search heuristic in \cref{alg:search}. The second is a policy extraction task, testing \cref{alg:imitation}, where the ground truth is a hand-coded policy. In these two first experiments, the DSL used for the search contains the true program used to generate observational data. Finally, in our third experiment we examine if we can learn a simple yet effective policy by imitation from a more complicated neural network policy which is trained using an existing reinforcement learning algorithm.

\subsection{Programming by example with local program search}
\begin{figure}[tb]
    \centering
    \includegraphics[scale=0.69]{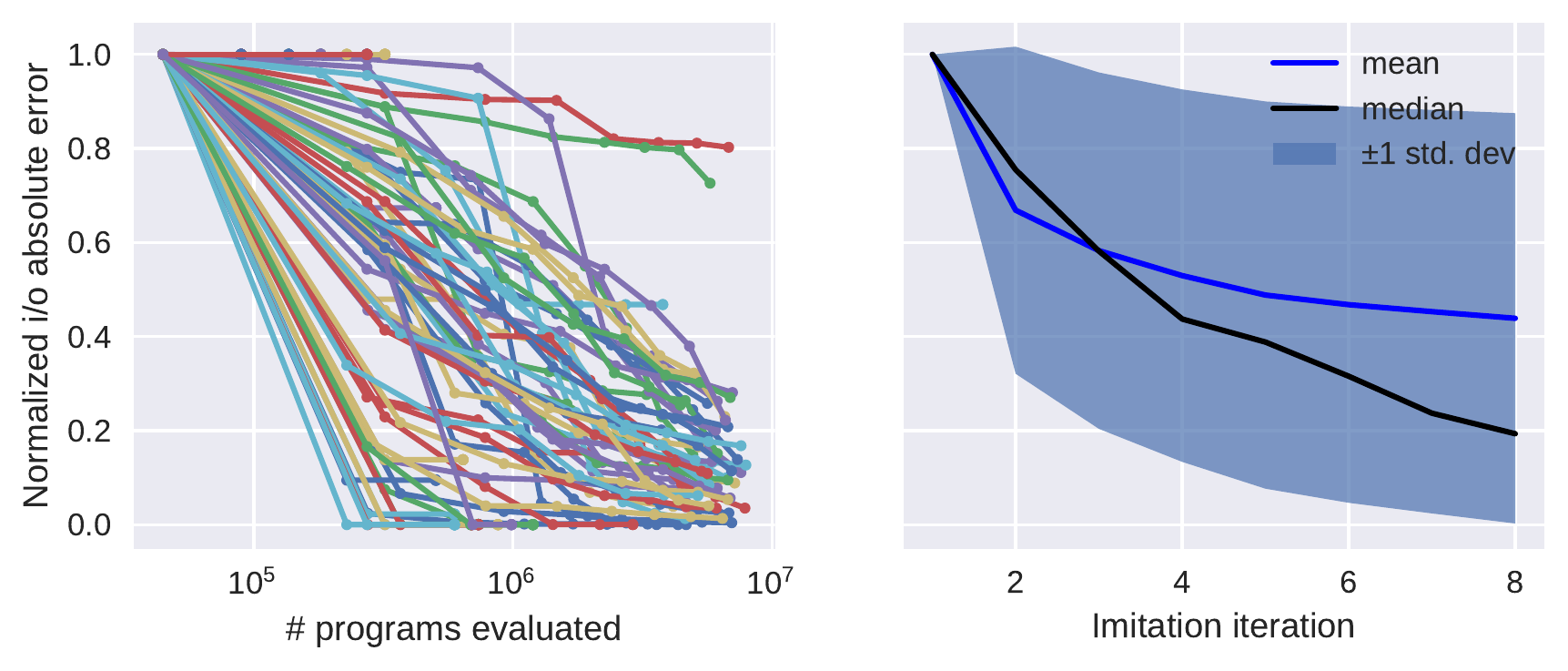}
    \caption{Programming by example: Learn a program from input-output examples. Experiment on 100 sampled programs. Left: For each program, the absolute error (normalized wrt. first iteration) by number of programs evaluated. Right: Mean, median, and standard deviation by search iteration.}
    \label{fig:random_programs}
\end{figure}
%  1. Introduction. What is the purpose of the experiment
%  2. Data, materials, methods. Exactly how was the experiment conducted.
%  3. Results
%  4. Discussion. Conclusions.
As a first evaluation of the method, we used a straightforward PBE task, whose purpose is to show that the described iterative local search is capable of synthesizing nontrivial programs from input-output specifications. 

\textbf{DSL:} We defined a language containing the set of constants $\{-1,\, 0,\, 0.5,\, 0.8,\,\allowbreak 1,\,\allowbreak 3,\,\allowbreak 5,\, 6,\, \true\}$, which are all $\Float$ing point numbers except $\true$ which is $\Bool$ean, and the functions with associated type signatures $\{\primif : \Bool \Rightarrow \T0 \Rightarrow \T0 \Rightarrow \T0,\, >\, : \Float \Rightarrow \Float \Rightarrow \Bool,\, \land\, : \Bool \Rightarrow \Bool \Rightarrow \Bool,\, \oplus : \Bool \Rightarrow \Bool \Rightarrow \Bool,\, - : \Float \Rightarrow \Float \Rightarrow \Float),\,\allowbreak * : \Float \Rightarrow \Float \Rightarrow \Float,\,\allowbreak \cdot^2\, : \Float \Rightarrow \Float\}$. 

\textbf{Data:} The observation space (i.e. input) to these programs consists of three $\Float$s, which are distinct variables that can be used just like constants. 10 sets of these numbers were randomly sampled as inputs to be used during synthesis. Ground truth programs of some length, as a simple proxy for complexity, were sampled from a weighted distribution over the DSL. In order to obtain samples that have a reasonable length, we designed a distribution on the abstract syntax of our DSL that puts more probability on higher-arity functions. Further, the probability of sampling $\true$ was weighted significantly down, while the probability of sampling an input variable was weighted higher. Since program length is not the best measure of complexity, samples were rejected on other criteria as well. Programs were discarded if: the length of the program (number of tokens) was less than 8, program output was constant, or an input-output equivalent program, on some randomly chosen inputs, existed within a depth 4 search of the DSL. 

\textbf{Results and discussion:}
Results from $d=4$ local search on 100 sampled ground truth programs can be seen in \cref{fig:random_programs}, which shows that for many of the programs an exact fit is found on the given inputs. Even for programs where an exact solution is not found, most of the searches show significant progress through the iterations, although a few make no progress at all. Since the search is deterministic, if no improvement is made in an iteration, further iterations will not lead to better results. It should also be noted that a single iteration of search with $d=5$ in this setting corresponds to evaluating about as many programs as 20 iterations with $d=4$.

\subsection{Imitation of a programmatic pendulum swing-up policy}
%  1. Introduction. What is the purpose of the experiment
%  2. Data, materials, methods. Exactly how was the experiment conducted.
% - description of pendulum swing-up environment + representation / input + what is the reward?
% - definition of DSL
% - definition of ground truth policy
% - definition of one iteration of imitation (search depth etc.)
% - what is the training data? (fixed set of trajectories + x expert labelled trajectories from the current programmatic policy
%  3. Results
%  4. Discussion. Conclusions.
% - why does the method not find the exact ground truth?

\newcommand{\rb}[1]{\raisebox{-0.85\height}{#1}}
\begin{figure}[htp]
    \centering
    \begin{tabular}{rcrc}
    & Imitation from {\bfseries programmatic} policy 
    & & Imitation from {\bfseries neural} policy
    \\
    a)& \rb{\includegraphics[width=0.47\textwidth]{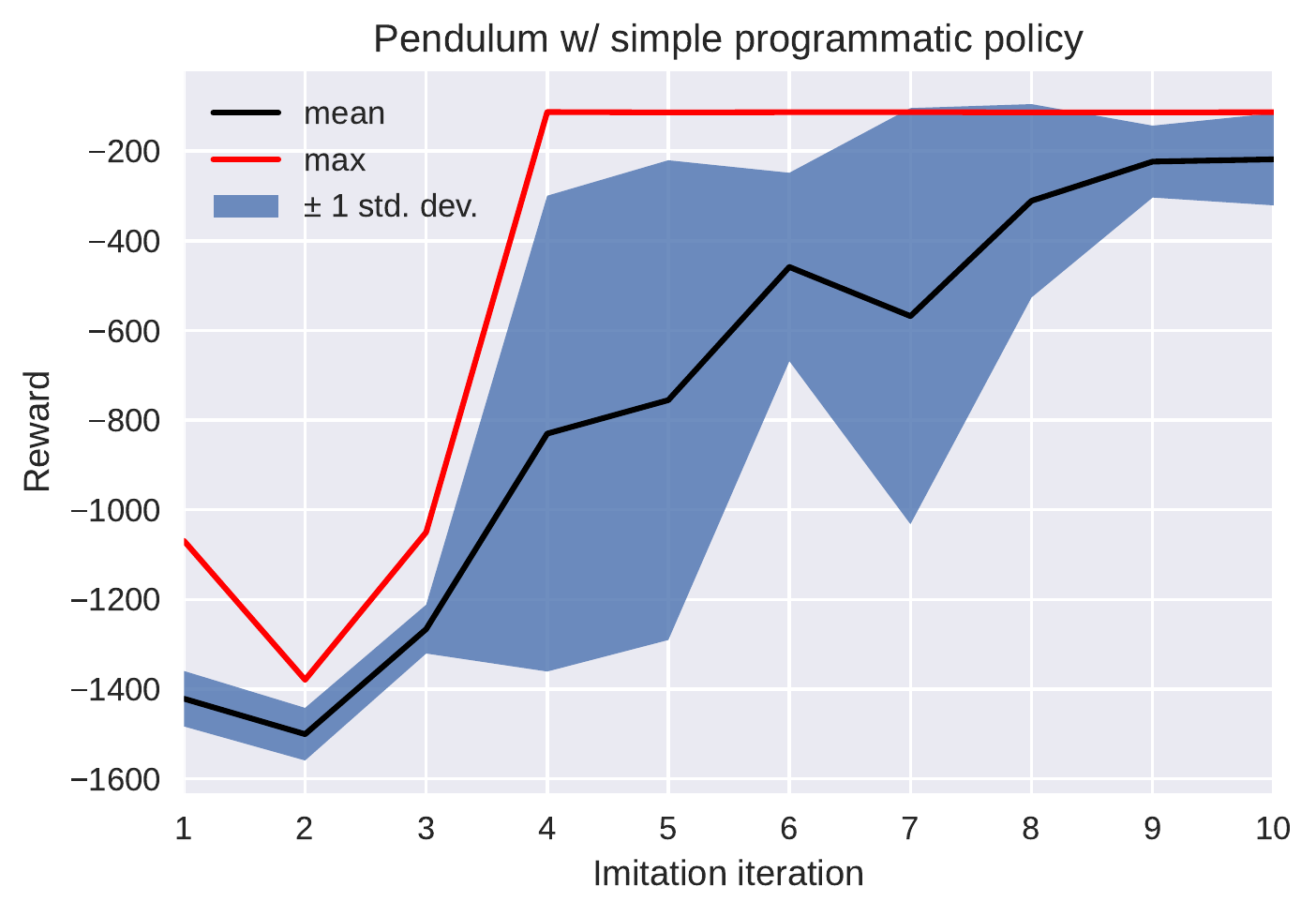}}
    &  & \rb{\includegraphics[width=0.47\textwidth]{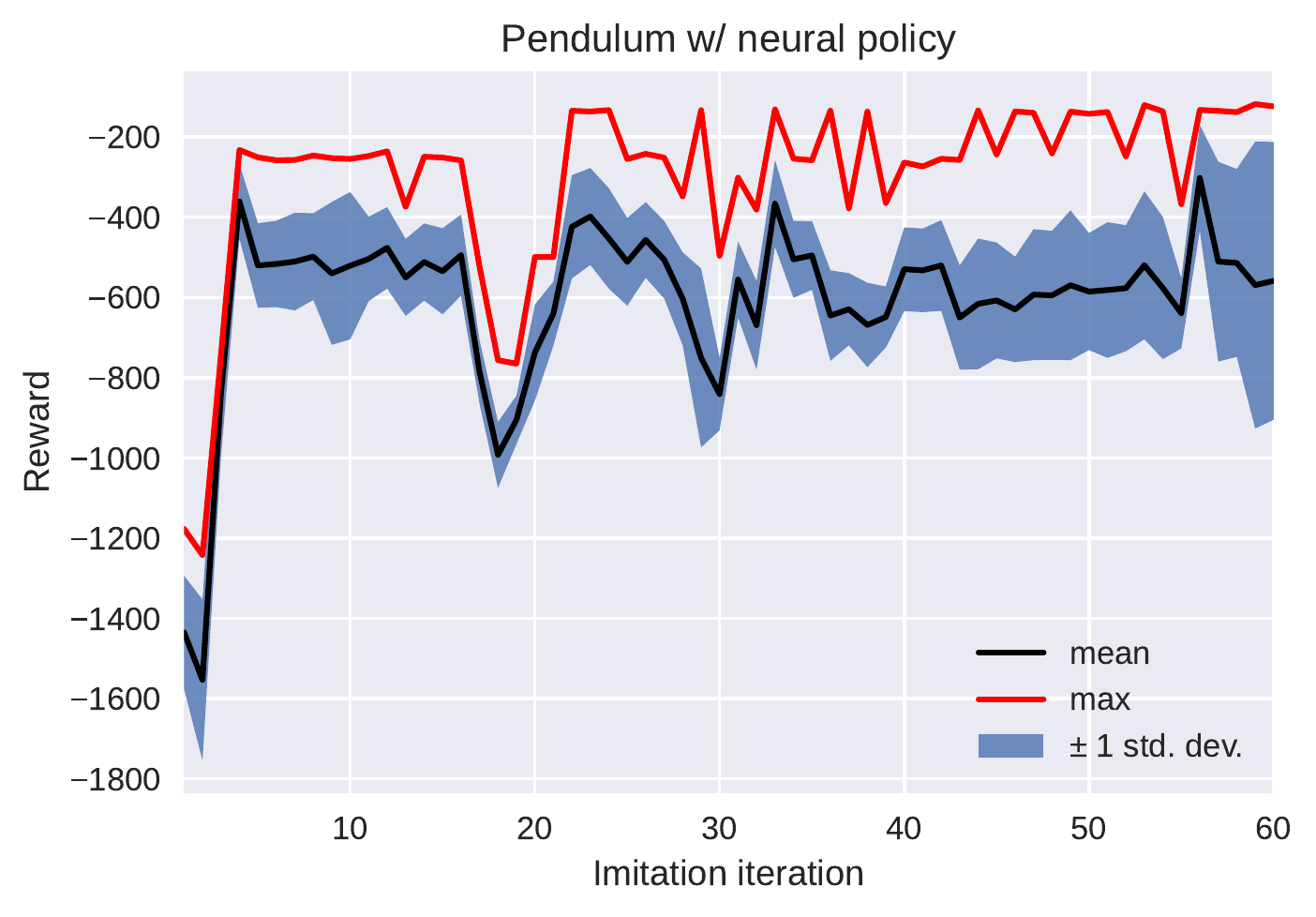}}
    \\
    b) & \rb{\includegraphics[width=0.47\textwidth]{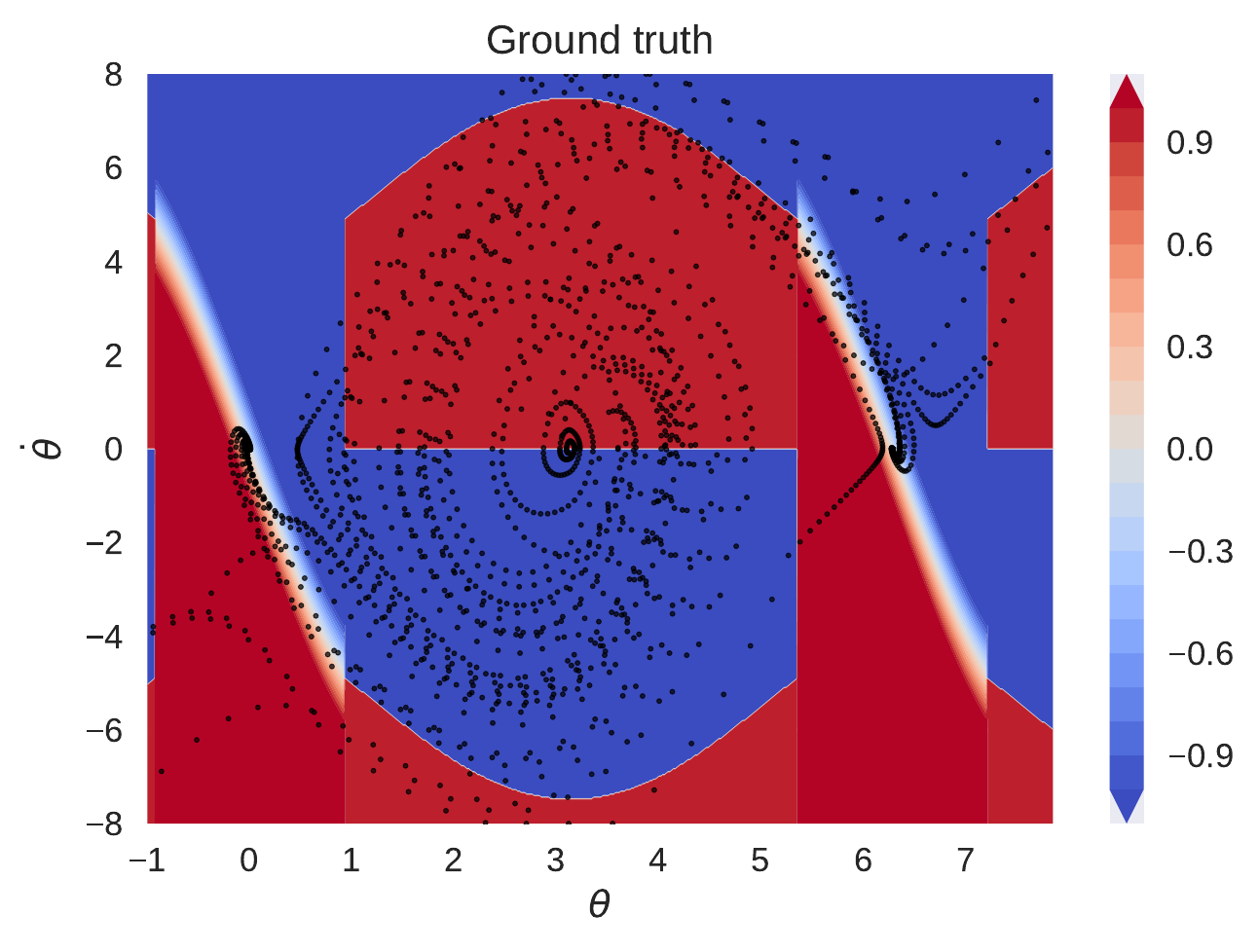}}
    &  & \rb{\includegraphics[width=0.47\textwidth]{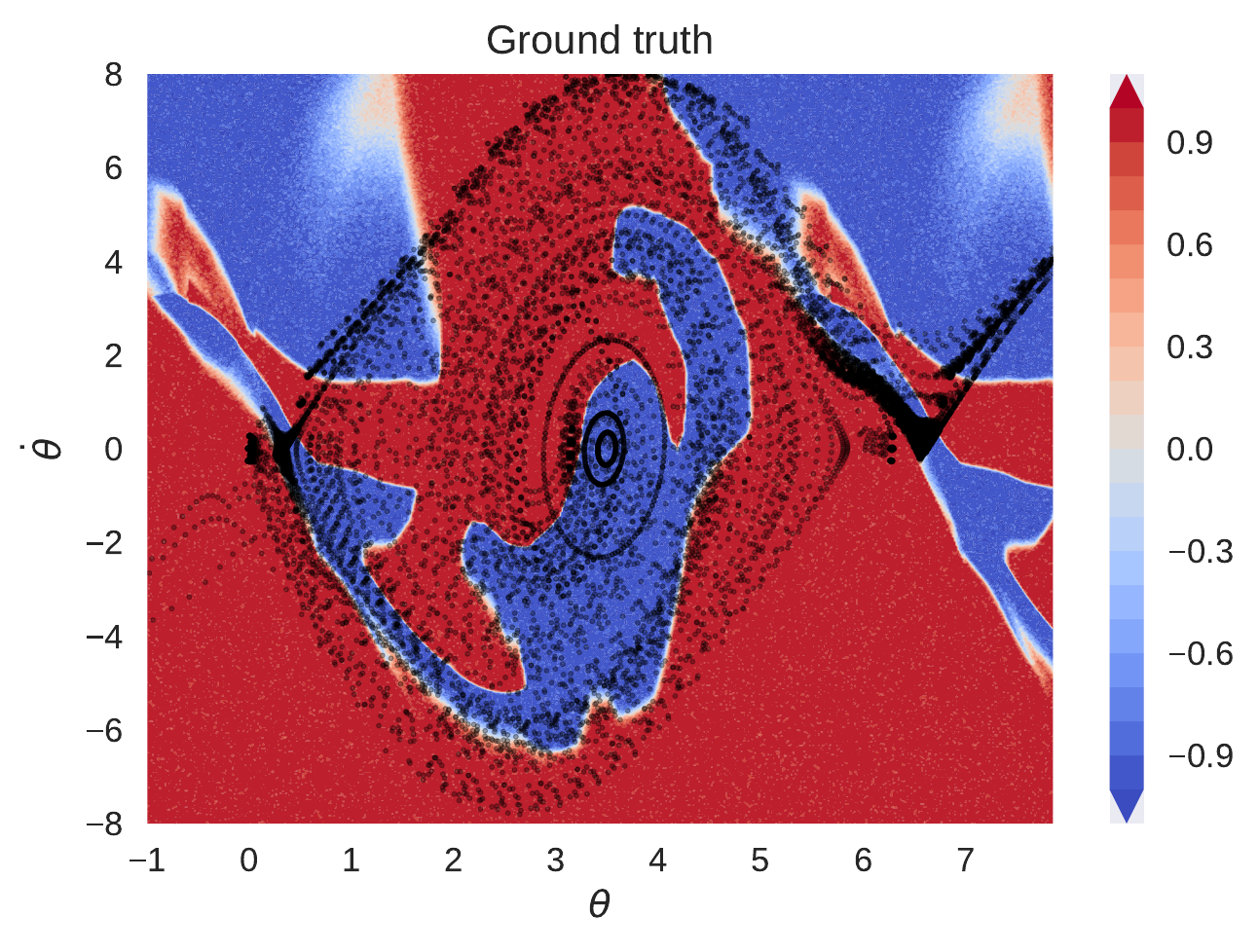}}
    \\
    c) & \rb{\includegraphics[width=0.47\textwidth]{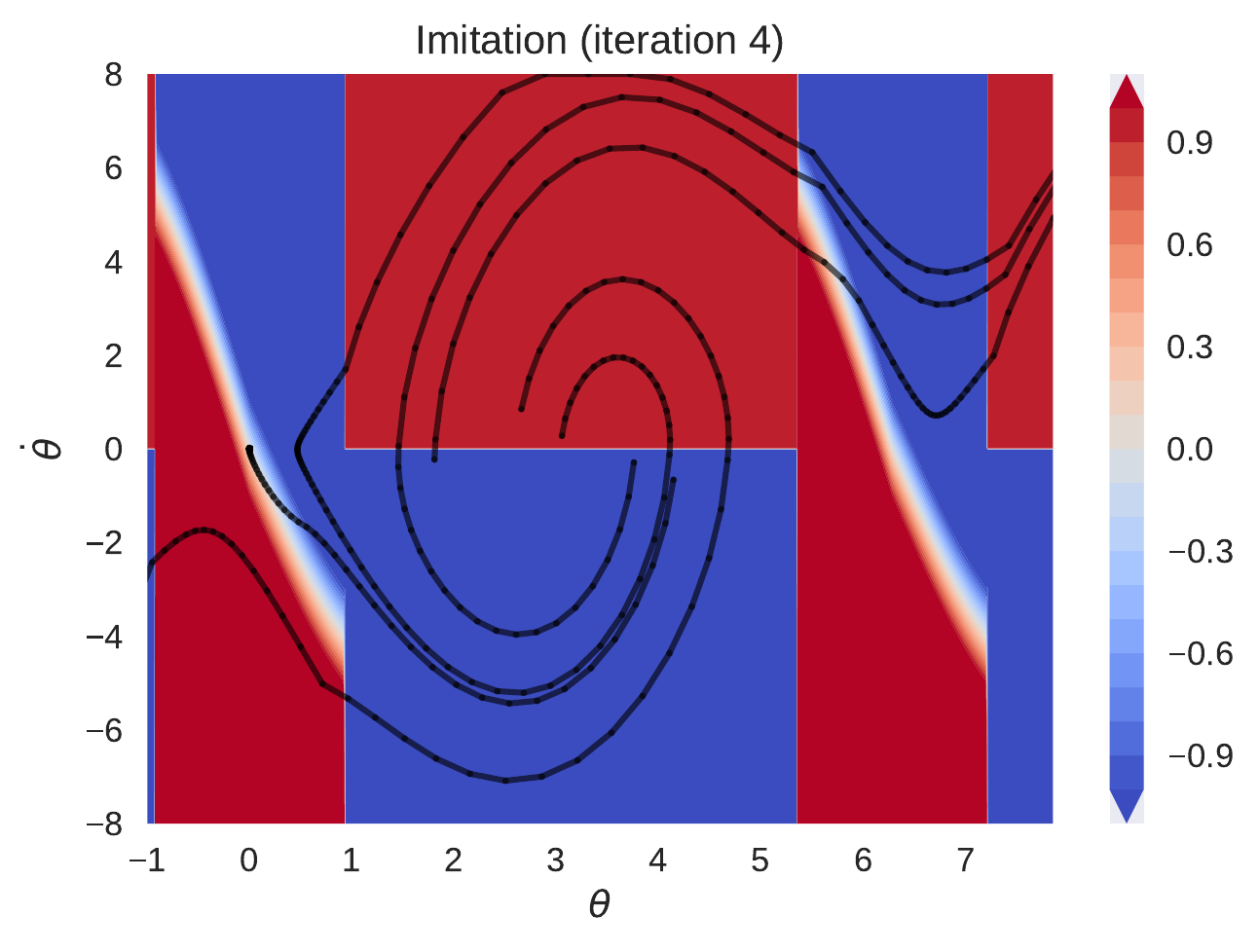}}
    &  & \rb{\includegraphics[width=0.47\textwidth]{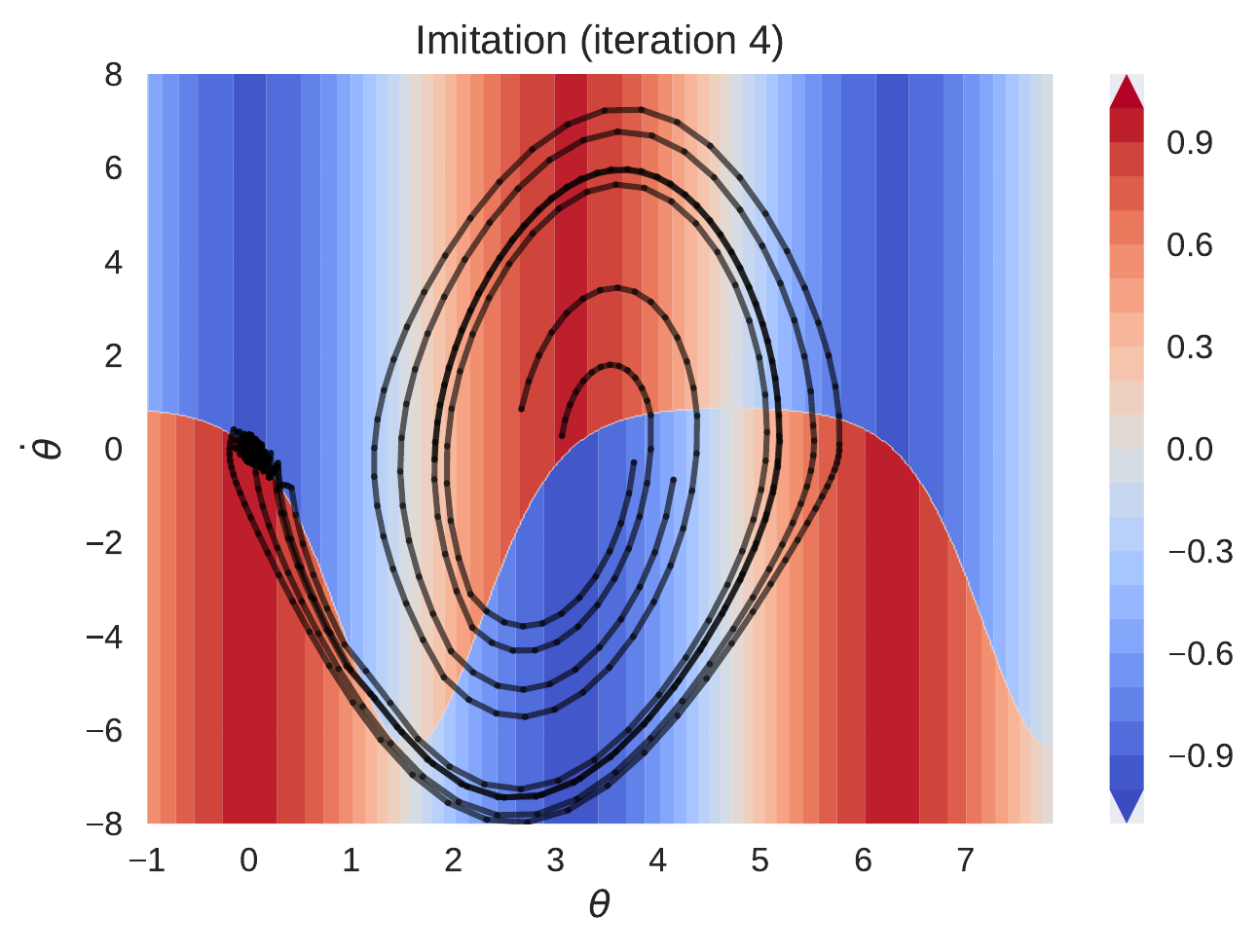}}
    \\
    d) & \rb{\includegraphics[width=0.47\textwidth]{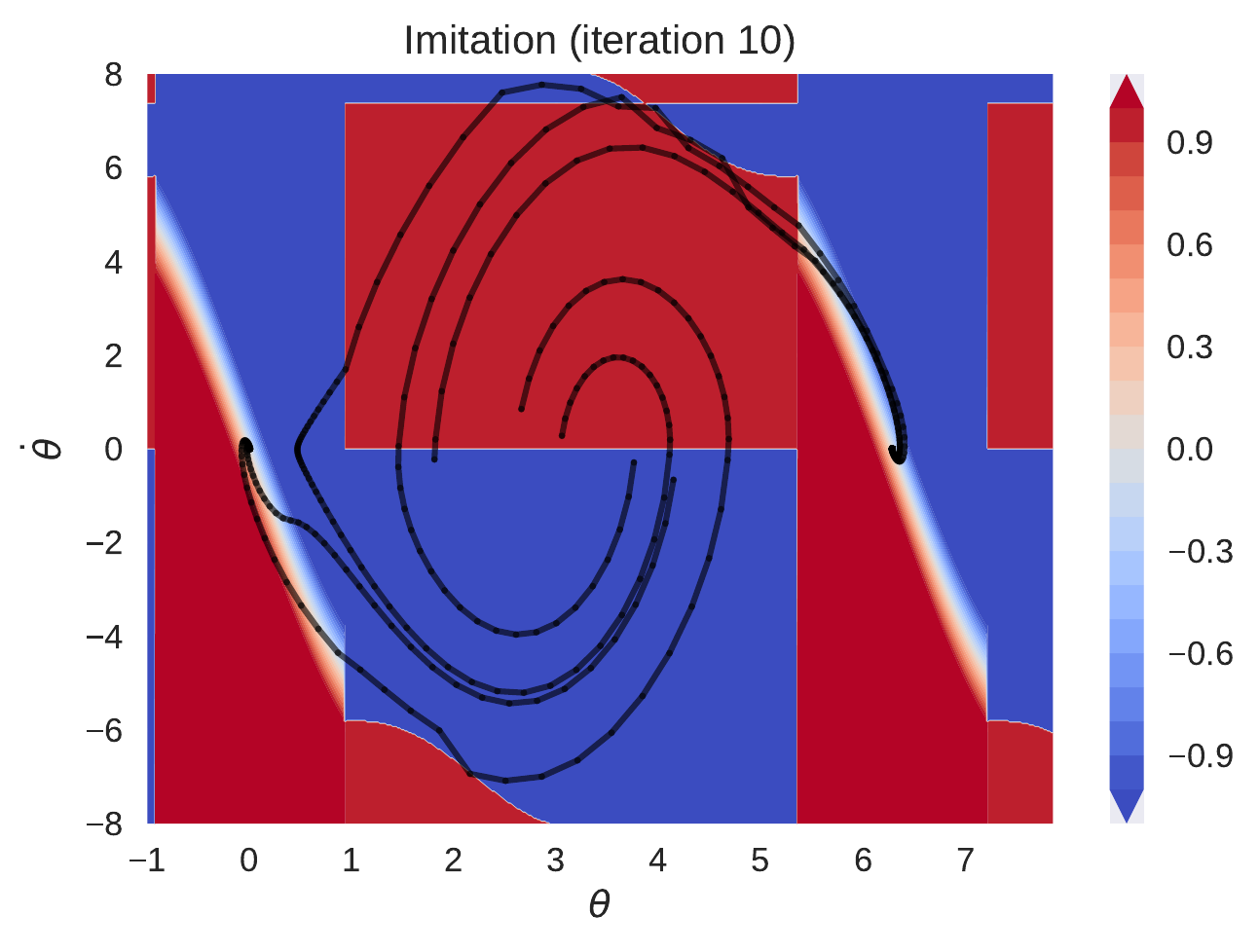}}
    &  & \rb{\includegraphics[width=0.47\textwidth]{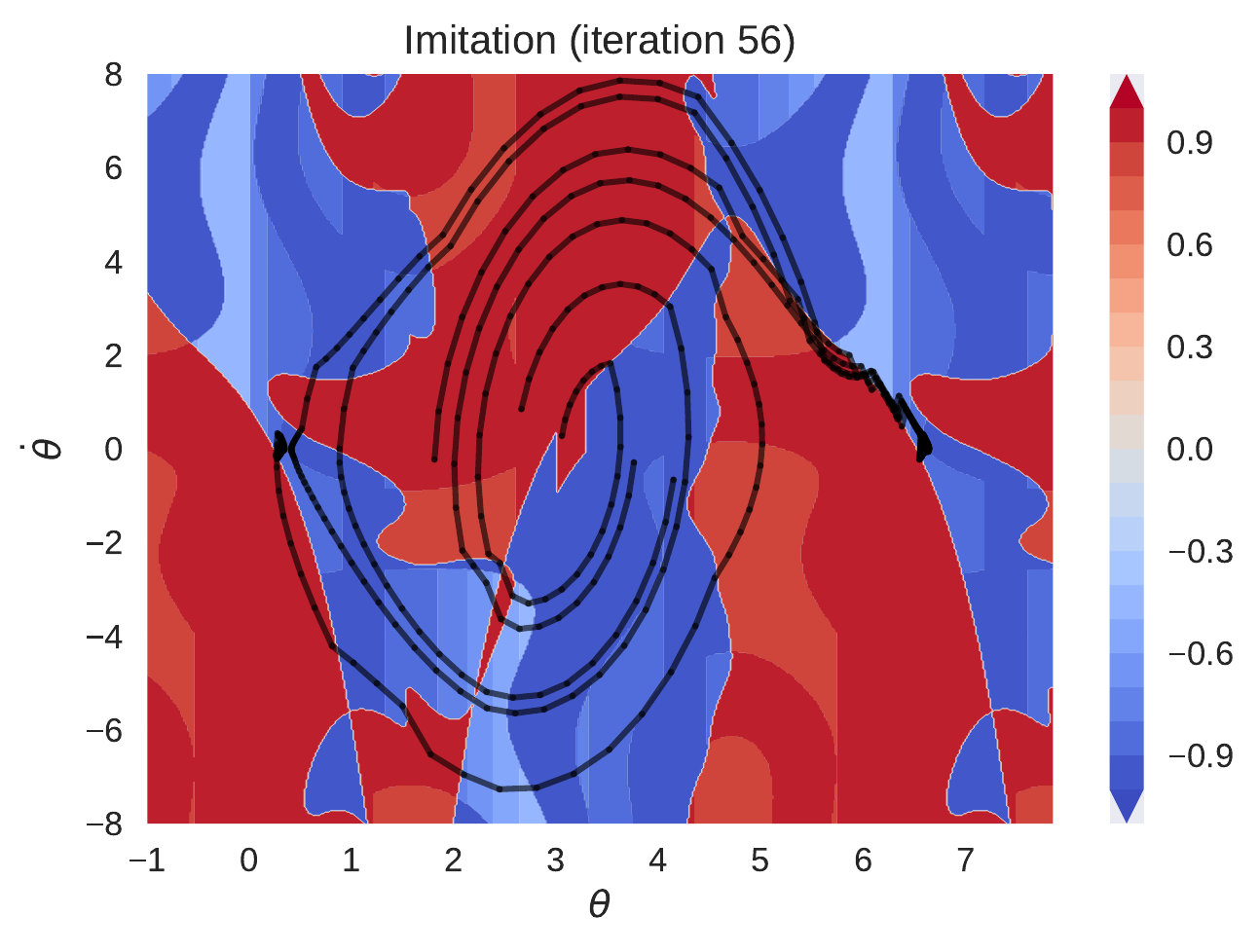}}
    \end{tabular}
    \caption{Pendulum swing-up imitation learning of a programmatic policy from a ground truth programmatic (left) or neural (right) policy. Policies are visualized as a heat map; the state space is pendulum angle, $\theta$, and angular velocity, $\dot\theta$, and the action is pendulum torque. The goal state is $\theta=0 \pmod{2\pi}$, $\dot\theta=0$. a) Cumulative reward of test trajectories. b) Ground truth programmatic/neural policy. Points indicate all states seen during training. c) Programmatic policy found after four iterations of imitation learning, with five test trajectories shown. d) Programmatic policy found after several more iterations, with five test trajectories shown.}
    \label{fig:pendulum}
\end{figure}

Next we examined if we were able to discover a ground truth programmatic policy. 

\textbf{Task:}
We based the experiment on a simple, classical control problem, the pendulum swing-up task. The state space consists of the angle and angular velocity of the pendulum, and the action space is the torque applied to the base of the pendulum, normalized to the interval [-1, 1]. This two-dimensional state space allows us to easily display and visually compare policies. The simulator is discretized with a time step of $0.05$s, and an episode is 200 steps long. The reward function is $r(\theta, \dot\theta, a) = ((\theta + \pi \pmod{2\pi}) - \pi)^2 + 0.1\dot\theta + 0.001a^2$, which results in a reward of $0$ if the pendulum is perfectly balanced with no torque applied, and a reward of $-\pi^2$ when the pendulum is pointing straight down while not moving. While the state space of the pendulum task is $(\theta, \dot\theta)$, the observation space supplied to the policies is $(\scalebox{0.85}{$\mathtt{x1, x2, x3}$}) = (\sin\theta, \cos\theta, \dot\theta)$.

\textbf{DSL:}
We used a simple, pure DSL with primitives suitable for solving the RL task, containing the constants $\{-6, -1, 1, 0.5, 0.6, 8, 10\}$, and the functions $\{\primif : \Bool \Rightarrow \T0 \Rightarrow \T0 \Rightarrow \T0,\, \footnotesize\texttt{gt}\, : \Float \Rightarrow \Float \Rightarrow \Bool,\, \footnotesize\texttt{sub} : \Float \Rightarrow \Float \Rightarrow \Float,\, \footnotesize\texttt{add} : \Float \Rightarrow \Float \Rightarrow \Float,\, \footnotesize\texttt{mul} : \Float \Rightarrow \Float \Rightarrow \Float,\, \sign : \Float \Rightarrow \Float,\, \footnotesize\texttt{sqr} : \Float \Rightarrow \Float\}$.

\textbf{Ground truth:}
We hand crafted a ground truth policy in the DSL, capable of swinging up the pendulum from any starting state. The policy achieves an average reward of approximately -211 and a maximum of -113. This handcrafted policy is given by the program {\footnotesize\texttt{((if ((gt x1) 0.6)) ((sub ((mul x2) -6)) x3)) (sign ((mul ((sub ((add ((mul 0.5) (sqr x3))) ((mul ten) ((sub x1) 1)))) \linebreak 8)) ((mul -1) x3)))}}.

\textbf{Synthesis:}
At each iteration of the local search we used a search depth of $d=4$ which was found to be enough to discover the $\primif$ expression that switches between swing-up and balancing. As training data we used $N=5$ state trajectories from the ground truth policy and $M=2$ trajectories from the latest programmatic imitation policy. All training states were expert labelled with actions from the ground truth policy.
For evaluation we simulated 100 rollouts from uniformly random states in the range $\tfrac{\pi}{2}\le\theta\le \tfrac{3\pi}{2}$, $-1\le\dot\theta\le 1$, which is the pendulum below horizontal with relatively low velocity.

\textbf{Results and discussion:}
The results of the experiment is shown in \cref{fig:pendulum} (left column). After four iterations of imitation learning a simple policy was found, capable of balancing the pendulum and swinging up from some states. After approximately ten iterations the policy could effectively swing up and balance the pendulum from any state. The imitation learning did not find the ground truth programmatic policy by iteration 10, likely due to the small number of observations in certain areas of the state space. Nonetheless, it managed to synthesize an effective policy which is quite similar to the ground truth.

\subsection{Imitation of a neural pendulum swing-up policy}
Finally, we examined if we were able to discover a simple, interpretable policy in a more realistic setting, with synthesis by imitating a trained neural policy. The task, DSL, and synthesis procedure were as described in the previous experiment, with the ground truth policy as the only difference.

\textbf{Ground truth:}
The neural ground truth policy was found by TD3 \cite{Fujimoto_van_Hoof_Meger_2018}, using feed-forward neural networks with 2 hidden layers of 24 neurons for both the actor and critic. Training was run for 5 million steps with a learning rate of $10^{-4}$ to ensure relatively good convergence.

\textbf{Results and discussion:}
The results of the experiment is shown in \cref{fig:pendulum} (right column). After four iterations of imitation learning, a simple imitating policy capable of swinging up and balancing the pendulum is found. This imitation policy is {\footnotesize\texttt{(mul x1) (cos (exp (sign ((add x3) ((add -1) (sqr (exp x2)))))))}}. After several more iterations, at iteration 56, a significantly more complicated programmatic policy was found which resembles the neural policy more closely but yields only a minor performance improvement, while being significantly less interpretable. This imitation policy has a length of 121 tokens, i.e. function calls plus arguments.

\section{Discussion}
We have presented and evaluated our method with simple experiments, and much remains to be done. As mentioned, one goal is to integrate the local search with reinforcement learning as described by \cref{alg:ip}. While simple, we believe that the presented results show potential, especially through the programs that were discovered in only a few iterations. In particular, it would be interesting to evaluate this approach on more structured tasks, where neural networks might struggle with generalization while a program could be found that immediately generalizes. In such a setting, we could also take better advantage of type-directed search, with more complicated DSLs containing e.g. logic, matrix or computer vision functions potentially still remaining computationally tractable.

It should also be mentioned that local, iterated synthesis as a concept remains orthogonal to several other improvements in program synthesis; for example, enumerating or sampling programs according to a learned probability distribution as in e.g. \cite{Ellis2018} is possible, as is better filtering as in \cite{Katayama_2008}. Instead of depth-limited search, it would be possible to limit the search to programs above a certain likelihood. However, it seems unclear how this distribution would be effectively learned for policies.
\subsection{Related work}
%By far the most related field is \emph{Evolutionary Computing}. Specifically, Genetic Programming is the application of methods from Evolutionary Computing to generating programs (cite). A basic Genetic Programming algorithm consists of a set of steps, usually called \emph{generations}. In each generation there is a \emph{population} of programs, which are \emph{selected} according to some fitness criterion. New programs are introduced by means of \emph{crossovers} or \emph{mutations}, respectively mixing programs from the population together and generating modifications to programs from the population. Genetic algorithms are stochastic processes, since operations such as crossover or mutation are performed randomly. The Genetic Programming literature is extensive, but we will discuss what we found to be the most relevant previous work.

Previous work on synthesizing programmatic policies at the intersection of RL and genetic programming (GP) include GPRL \cite{Hein_Udluft_Runkler_2017} which is based on offline GP, performed in a previously learned parametric model of the system of interest. They include a comparison with behavioral cloning, i.e., direct imitation of the actions of a trained policy. Their method performs better on the actual (simulated, but not learned) system. It is well known that behavioral cloning can lead to poor performance, e.g. \cite{Ross2010-zq}, which could explain the observed performance gap. It seems likely that interaction with the model can overcome some of the distributional issues arising from behavioral cloning.
In RL, it might be preferable to not learn a parametric model if it is used for credit assignment (i.e. policy learning) \cite{van_Hasselt_Hessel_Aslanides_2019}. 
\cite{Gupta_Christensen_Chen_Song_2020} proposed a method for using program repair in neural program synthesis. After neural synthesis, the resulting program might not be correct or even satisfy the input-output relation. The authors propose to learn a neural debugger that outputs so-called edits which correct potential errors in the program. The relation to this work is apparent in how we use an edit operator to define the neighborhood of a program.
\cite{Kamio_Mitsuhashi_Iba_2003} describe a way to integrate GP, RL and simulated systems. By first synthesizing a policy using GP in the simulated system, it can later be adapted and fine-tuned through RL, allowing the policy to function on a real robot.
\cite{Inala2020} describe an imitation learning method that improves the inductive generalization by adapting the teacher distribution according to the imitating policy.
The presented neighborhood search method can be considered an instance of the (Very) Large-Scale Neighborhood Search framework \cite{Pisinger_Ropke_2010}. Deterministic versions of genetic algorithms have been considered before, such as in \cite{Salomon_2003}.

%\newpage

% New references:
% \cite{Inala2020} adaptive teacher
% \cite{Yang2021} synthesize programs to guide reinforcement learning
% \cite{Srivastava2011} generalized planning
% \cite{Aguas2021} native heuristic search for generalized planning

%\bibliographystyle{apalike}
\urlstyle{rm}% modify as appropriate
\bibliographystyle{splncs04}
\bibliography{main}
%\printbibliography

\end{document}